\documentclass[runningheads]{llncs}
\usepackage{graphicx}

\usepackage{tikz}
\usepackage{comment}
\usepackage{amsmath,amssymb} 
\usepackage{color}

\usepackage[accsupp]{axessibility}  
\usepackage[misc]{ifsym} 

\begin{document}
\pagestyle{headings}
\mainmatter
\def\ECCVSubNumber{3580}  

\title{PointCLM: A Contrastive Learning-based Framework for Multi-instance Point Cloud Registration} 




\titlerunning{PointCLM}
%
\author{Mingzhi Yuan\inst{1,3} \and
Zhihao Li\inst{1,3} \and
Qiuye Jin\inst{1,3} \and \\
Xinrong Chen\inst{1,2,3}$^{\textrm{\Letter}}$ \and
Manning Wang\inst{1,3}$^{\textrm{\Letter}}$  }
\authorrunning{M. Yuan al.}
%
\institute{Digital Medical Research Center, School of Basic Medical Sciences, Fudan University, Shanghai 200032, China \and
Academy for Engineering and Technology, Fudan University, Shanghai 200032, China \and
Shanghai Key Laboratory of Medical Image Computing and Computer Assisted Intervention, Shanghai 200032, China \\
\email{\{mzyuan20,lizhihao21,qyjin18,chenxinrong,mnwang\}@fudan.edu.cn}}

\maketitle

\begin{abstract}
  Multi-instance point cloud registration is the problem of estimating multiple poses of source point cloud instances within a target point cloud. 
  Solving this problem is challenging since inlier correspondences of one instance constitute outliers of all the other instances. 
  Existing methods often rely on time-consuming hypothesis sampling or features leveraging spatial consistency, resulting in limited performance. 
  In this paper, we propose PointCLM, a contrastive learning-based framework for mutli-instance point cloud registration. 
  We first utilize contrastive learning to learn well-distributed deep representations for the input putative correspondences. 
  Then based on these representations, we propose a outlier pruning strategy and a clustering strategy to efficiently remove outliers and assign the remaining correspondences to correct instances. 
  Our method outperforms the state-of-the-art methods on both synthetic and real datasets by a large margin. 
  The code will be made publicly available at \url{http://github.com/phdymz/PointCLM}.
  
\end{abstract}

\section{Introduction}
\begin{figure}[htp]
  \centering
  \includegraphics[width=0.8\textwidth]{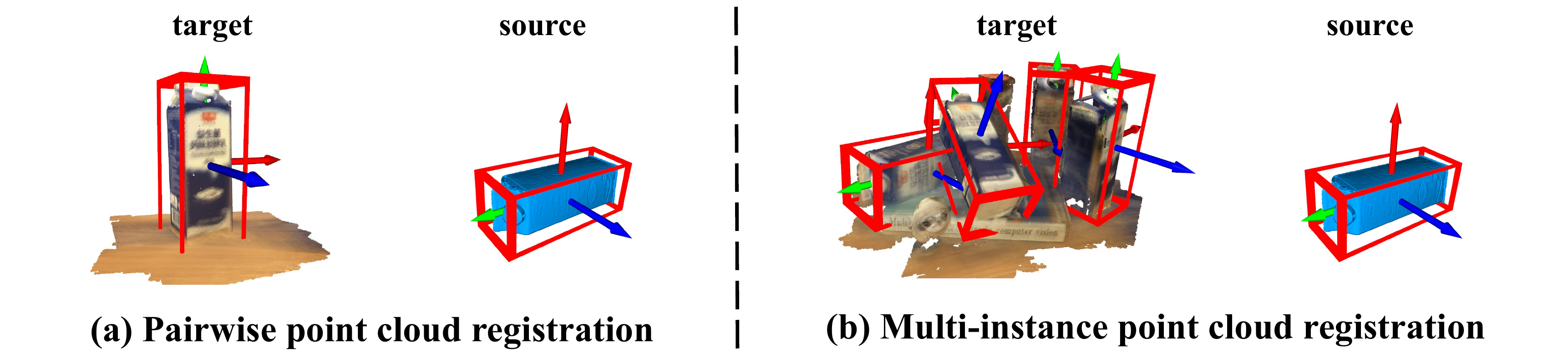}
  \caption{Given a source point cloud of a 3d object, pairwise point cloud registration (left) focuses on estimating a single rigid transformation between the source point cloud and the target point cloud, while multi-instance point cloud registration (right) aims to estimate the 6d poses of the same objects within the target point cloud.}
  \label{fig1}
\end{figure}

3D point cloud registration is a fundamental task in computer vision \cite{ref1,ref2,ref3}, and most studies mainly focus on pairwise registration. 
However, in real applications, target scene may contain multiple repeated instances, and we need to estimate multiple rigid transformations between a source point cloud and these repeated instances in the target point cloud. 
An example is illustrated in Figure \ref{fig1}. 
This problem is named as multi-instance point cloud registration and it is more challenging than pairwise point cloud registration. 

There exist two solutions to the multi-instance point cloud registration problem. 
One solution is to use an instance detector \cite{ref5,ref6} to detect instances in the target point cloud, and turn this problem into multiple pairwise registrations. 
However, this approach can only detect known classes in the training set, and the registration performance is limited by the instance detector. 
Another solution is via multi-model fitting \cite{ref4,ref14,ref42,ref43}. 
This approach starts from building putative correspondences based on local features, followed by estimating multiple transformations from noisy correspondences using multi-model fitting algorithms. 
Multi-model fitting problem has been studied for decades. 
The basic idea of most multi-model fitting methods is to sample a series of hypotheses and then perform preference analysis \cite{ref9,ref10,ref11,ref12} or consensus analysis \cite{ref13,ref14,ref15,ref16}. 
Most of these multi-model fitting algorithms are traditional methods, which rely on a large number of sampling for generating hypotheses and sophisticated strategies for selecting real models, resulting in large computational cost. 

Recently, a series of deep learning-based methods have achieved excellent performance in pairwise point cloud registration, including feature matching \cite{ref23,ref24,ref25,ref26}, outlier correspondences rejection \cite{ref17,ref18,ref19,ref20}, etc. 
Multi-instance point cloud registration also faces the interference of outliers, which inspires researchers to extend deep learning-based methods to the multi-instance case. 
However, in the multi-instance case, the inliers of one instance constitute outliers of all the other instances, so we need not only to identify outliers, but also to predict which instance the inliers belong to. 
To the best of our knowledge, the only existing deep learning-based method \cite{ref42} that can be used in multi-instance point cloud registration is by sequential binary classification. 
However, this method has limited performance because it does not fully explore the interaction between correspondences. 
There is also a traditional method \cite{ref4} which clusters the correspondences based on spatial consistency \cite{ref37}. 
However, due to the ambiguity of spatial consistency, this method cannot assign the correspondences very efficiently when outliers are located close to inliers. 
Therefore, an intuitive idea is to learn a more discriminative representation for correspondence so that not only the outliers can be easily pruned but also the inliers of different instances are separable in the feature space. 

In this paper, we propose a contrastive learning-based method to learn deep representations for correspondences, based on which, we can not only remove outliers in the putative correspondences, but also correctly assign inliers to each instance. 
Specifically, we first select inlier correspondences belonging to the same instance to build positive pairs and select one correspondence from an instance and another correspondence not belonging to that instance to build negative pairs. 
Then we train a feature extractor with contrastive loss to make the correspondences in positive pairs near each other and the correspondences in negative pairs fall apart from each other in the feature space. 
This makes the inliers of each instance form a certain scale of clusters in the feature space, while the outliers are scattered. Based on this distribution, we prune the input putative correspondences according to the density over the feature space and obtain a correspondence set that has little outliers. 
After that, we perform a spectral clustering based on feature similarity and spatial consistency \cite{ref37,ref19}, in which the number of instances can be automatically determined and the remaining correspondences can be correctly assigned to each instance. 
Finally, we calculate the transformation for each instance using the correspondences assigned to it. Our main contributions are as follows:
\begin{itemize}
  \item We propose a contrastive learning-based strategy that makes inlier correspondences and outlier correspondences well distributed in the feature space;
  \item We propose a pruning strategy based on feature similarity and spatial consistency to remove outliers and then utilize spectral clustering to assign the remaining correspondences to correct instances;
  \item Our method outperforms existing state-of-the-art methods in multi-instance point cloud registration on both synthetic and real datasets. 
\end{itemize}

\section{Related Work}
\textbf{Point cloud registration} has long been a fundamental task in computer vision and robotic, which can be roughly divided into direct methods \cite{ref2,ref55,ref56,ref57} and feature-based methods \cite{ref24,ref25,ref26,ref62}. 
In recent years, thanks to the development of deep learning, many feature-based methods achieved state-of-the-art performance. 
These methods commonly produce correspondences by feature matching and then remove outliers to estimate transformations robustly. 
Despite the rapid development of deep features \cite{ref23,ref24,ref25,ref26}, the correspondences generated by feature matching still contain outliers. 
Therefore, removing outliers is of great significance in point cloud registration. 
In the past, many traditional methods have been proposed to remove outliers, including RANSAC-based methods \cite{ref9,ref27,ref28}, branch and bound-based methods \cite{ref30}, and many others \cite{ref31,ref32}. 
A comprehensive review of these methods can be found in \cite{ref1,ref3,ref33}. 
Recently, a series of learning-based methods \cite{ref19,ref20,ref17,ref18} have been proposed and achieved remarkable results in outlier removal. 
For example, Yi et al. used a PointNet-style \cite{ref34} network with instance normalization to predict outliers \cite{ref17}, which has been widely used as a backbone network for predicting outliers \cite{ref35,ref36}. 
Choy et al. used a sparse convolution-based network to classify putative correspondences into inliers and outliers \cite{ref18}. 
Based on the assumption of spatial compatibility between inliers, Bai et al. incorporated geometric prior into deep neural network and designed a non-local layer \cite{ref19} to better aggregate features, which achieved outstanding results. 
The above methods are all designed for pairwise registration. 
However, unlike pairwise registration, inliers of one instance constitute outliers of all the other instances in multi-instance point cloud registration. 
Such pseudo outliers make it difficult to directly generalize the above binary classification models to the case of multi-instance point cloud registration. 
Inspired by success of previous works in pairwise registration, we introduce deep learning into multi-instance point cloud registration and propose a method that can not only remove outliers but also assign inlier correspondences to each instance. 

\noindent
\textbf{Multi-model fitting} aims to fit multiple models from noisy data, such as fitting multiple planes \cite{ref59} in a point cloud, estimating fundamental matrices in motion segmentation \cite{ref38}, calculating rigid transformations in multi-instance point cloud registration \cite{ref4}, etc. 
Since inliers of one instance constitute outliers of all the other instances, multi-model fitting is more challenging than single-model fitting. 
Existing multi-model fitting methods can be roughly divided into two categories. 
The first category fits models sequentially \cite{ref39,ref40,ref41,ref42}, which relies on repeatedly sampling and selecting models. 
For example, sequential RANSAC \cite{ref39} detects instances in a sequential manner by repeatedly running RANSAC to recover a single instance and then removing its inliers from the input. 
Progressive-X and Progressive-X+ \cite{ref40,ref41} use a better performing Graph-cut RANSAC \cite{ref27} as a sampler to generate hypotheses. 
CONSAC \cite{ref42} introduced deep models into multi-model fitting for the first time, using a network similar to PointNet \cite{ref34} to guide sampling. 
The second category fits multiple models simultaneously \cite{ref4,ref13,ref14,ref15,ref43}. 
For example, many preference analysis-based methods \cite{ref13,ref14,ref15} initially sample a series of hypotheses and then cluster input points according to the residuals of the hypotheses. 
RansaCov \cite{ref43} formulates the multi-model fitting as a maximum coverage problem and provides two strategies to solve it approximately. 
ECC \cite{ref4} utilizes the spatial consistency \cite{ref37} of point cloud rigid transformation and clusters correspondences in a bottom-up manner based on a distance-invariant matrix. 
Although spatial consistency performs efficiently in \cite{ref4}, the lack of orientation constraints makes the distance-invariant matrix still ambiguous in some cases, especially when outliers are close to inliers. 
In this paper, a novel deep representation is integrated with the spatial consistency to achieve better results.

\section{Problem Formulation}
We use $X$ and $Y$ to denote the source and target point clouds, respectively. 
The source point cloud consists of an instance of a 3D model, and the target point cloud contains $M$ instances of the same model, where these instances may be sampled from a part of the 3D model. 
By matching local features \cite{ref23,ref24,ref25,ref26}, we can generate putative correspondences between the two point clouds. 
A correspondence is denoted $c_{i}=\left(x_{i}, y_{i}\right) \in \mathbb{R}^{6}$, where $x_{i} \in X$, $y_{i} \in Y$ are the coordinates of a pair of 3D keypoints from the two point clouds. 
Our objective is to divide the putative correspondence set $C=\left\{c_{i}\right\}_{i=1}^{N}$ into $M+1$ subsets $C_{o} , C_{1} , \ldots , C_{M}$ satisfying $C=C_{o} \cup C_{1} \cup \ldots \cup C_{M}$, where $C_o$ denotes the predicted outlier set and $C_m$ denotes the inlier set for the $m$-th predicted instance. 
When we know the true instance number is $M$, recovering $M$ rigid transformations $\left\{R_{m}, t_{m}\right\}_{m=1}^{M}$ from the two point clouds is to minimize the objective function:
\begin{equation}
  \min _{\left\{R_{m}, t_{m}\right\}_{m=1}^{M}} \frac{1}{M} \sum_{m=1}^{M} \sum_{\left(x_{m i}, y_{m i}\right) \in C_{m}^{g t}} \frac{1}{\left|C_{m}^{g t}\right|}\left\|y_{m i}-R_{m} x_{m i}-t_{m}\right\|^{2}
\end{equation}
where $C_{m}^{g t}$ denotes the ground truth inlier set of the $m$-th instance, and ${\left|C_{m}^{g t}\right|}$ denotes the number of inliers in $C_{m}^{g t}$. 
The above problem is very challenging, because the inliers in $C_{i}^{g t}$ constitute outliers in $C_{j}^{g t}$ for $i \neq j$. 
In practice, the problem becomes even more difficult because the true instance number in the target point cloud is often unknown in prior, which is the case that we deal with in this paper.

\begin{figure}[ht]
  \centering
  \includegraphics[width=1.0\textwidth]{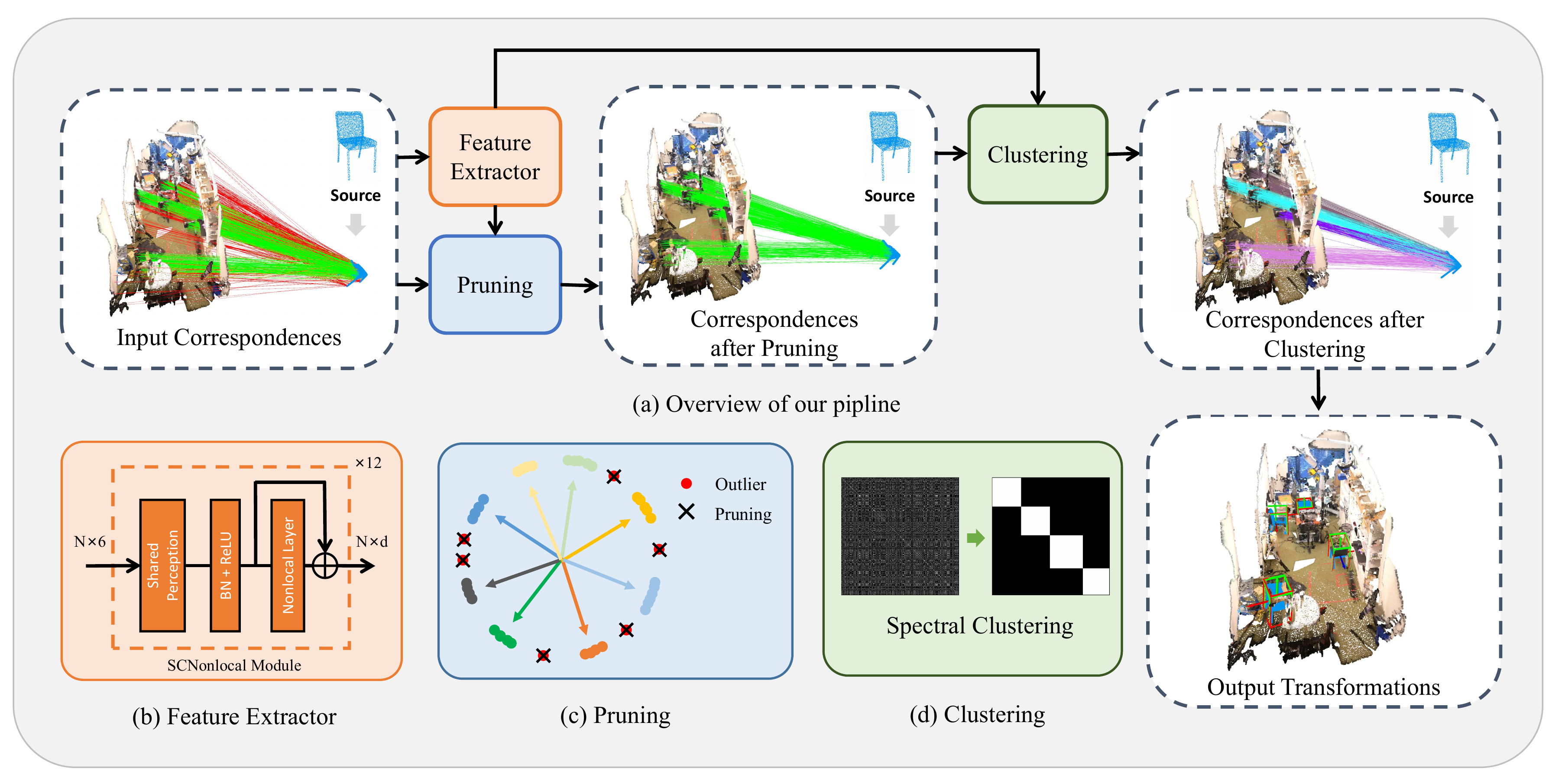}
  \caption{
    The pipeline of the proposed multi-instance point cloud registration framework PointCLM. 
    It takes putative correspondence as input, and output $M$ rigid transformations. 
    The green lines and red lines represent inliers and outliers, respectively. 
    After clustering, the correspondences in different clusters are visualized in different colors. 
    The green bounding boxes in output transformations represent the ground truth poses of instances in the target point cloud and the red bounding boxes represent our predictions. 
    The transformed point clouds in the target point cloud are visualized in blue.
  }
  \label{fig2}
\end{figure}

\section{Method}
In this section, we present our framework for multi-instance point cloud registration, which is illustrated in Figure \ref{fig2}. 
Our framework takes the putative correspondences generated by feature matching as input, and it first uses the Feature Extractor trained by contrastive learning to extract deep representations for the input correspondences (section \ref{sec4.1}). 
Then, we prune the correspondences according to both spatial consistency and the similarity of their deep representations (section \ref{sec4.2}). 
Finally, we cluster the remaining correspondences and estimate the transformations for multiple instances using the clustering results (section \ref{sec4.3}). 

\subsection{Feature Extractor}\label{sec4.1}
The first stage of our framework embeds the $N$ input putative correspondences $C=\left\{c_{i}\right\}_{i=1}^{N}$ into a feature space to obtain well-distributed $d$-dimension representations $F=\left\{f_{i} \in \mathbb{R}^{d}\right\}_{i=1}^{N}$ for the following pruning and clustering. 
Here, we adopt the SCNonlocal module in \cite{ref19} as our feature extractor, which consists of 12 repetitive blocks. 
As shown in Figure \ref{fig2}(b), each block consists of a shared Perceptron layer, a BatchNorm layer with ReLU and a nonlocal layer. The nonlocal layer integrates the spatial consistency \cite{ref37} of rigid transformation. 
Before calculating the features of each layer, a spatial consistency matrix $\beta$ is first calculated:
\begin{equation}
  \beta_{i j}=\left[1-\frac{d_{i j}^{2}}{\sigma_{d}^{2}}\right]_{+}, d_{i j}=\left|\left\|x_{i}-x_{j}\right\|-\left\|y_{i}-y_{j}\right\|\right|
\end{equation}
where $\sigma_{d}$ is a distance parameter to control the sensitivity to length difference, $[.]_+$ denotes a clamp function $max(x,0)$ to make $\beta_{ij} \geq 0$. 
The nonlocal layers aggregate the intermediate features using $\beta$:
\begin{equation}
  f_{i}^{k+1}=f_{i}^{k}+M L P\left(\sum_{j=1}^{|C|} \operatorname{soft} \max _{j}(\alpha \beta) g\left(f_{j}^{k}\right)\right)
\end{equation}
where $\alpha$ denotes the embed dot-product similarity between the intermediate feature representations $f_i^k$ and $f_j^k$ in the $k$-th blocks, and $g(\cdot)$ denotes a linear projection function. 
More details about the network can be found in \cite{ref19}.

We utilize contrastive learning to train our feature extractor to obtain well-distributed deep representations $F=\left\{f_{i} \in \mathbb{R}^{d}\right\}_{i=1}^{N}$. 
Concretely, for an anchor correspondence $c_{m i} \in C_{m}^{g t}$, we define the other correspondences in $C_m^{gt}$ as its positive samples, and define the correspondences in $C \backslash C_{m}^{g t}$ as its negative samples. 
We define the negative sample with the smallest Euclidean distance to the anchor correspondence in the feature space as the hardest negative sample. 
During each iteration, we exhaust all inliers as anchors and select their hardest negative samples to build hardest negative pair set $\mathcal{N}$ and randomly select positive samples to build positive pair set $\mathcal{P}$. 
Our contrastive loss is formulated as:
\begin{equation}
  \mathcal{L}=\sum_{(i, j) \in \mathcal{P}}\left[D\left(f_{i}, f_{j}\right)-m_{p}\right]_{+}^{2} /|\mathcal{P}|+\sum_{(i, j) \in \mathcal{N}}\left[m_{n}-D\left(f_{i}, f_{j}\right)\right]_{+}^{2} /|\mathcal{N}|
  \label{loss}
\end{equation}
where $m_p$ and $m_n$ are margins for positive and negative pairs, which prevent the network from overfitting \cite{ref58}. 
Since the inlier rate of input correspondences is commonly not high and we can downsample the input correspondences to a suitable scale, our exhaustion over all negative samples to find the hardest one for each minibatch is feasible. 
Our ablation experiments show that utilizing hardest negative pairs plays an important role in the high performance of the proposed method. 

By optimizing the loss in equation \ref{loss}, the representations of correspondences belonging to different instances are easily separable in the feature space, while the representations of correspondences belonging to the same instance are clustered together. 
We then use these well-distributed representations for both pruning and clustering. 
Our experiments show that the deep representations provide information that spatial consistency does not have, which makes pruning and clustering more effective. 

\subsection{Pruning}\label{sec4.2}
The input correspondences tend to contain a large proportion of outliers, which severely devastates the following inlier correspondence clustering and transformation estimation. 
Therefore, an intuitive idea is to first prune the input correspondences to remove outliers. 
A series of pruning strategies have been proposed in the past. 
For example, the input can be pruned based on similarity threshold \cite{ref60} or confidence \cite{ref35}. 
In this paper, we propose a density-based pruning strategy which prunes the input correspondences according to the deep representation learned in Section \ref{sec4.1} along with spatial consistency constraints.

As introduced in Section \ref{sec4.1}, through contrastive learning, correspondences satisfying the same rigid transformation have similar representations in the feature space and they are far away from other correspondences. 
This means that the feature density around inliers are higher than that around outliers \cite{ref61}. 
Based on this idea, we design a density-based pruning strategy as shown in Figure \ref{fig2}(c). 
We treat the isolated points in red as outliers and remove it, while the clustered points in other colors are reserved. The detail of our pruning method is as follows. 

We first calculate the feature similarity matrix $S^F$ between correspondences using the extracted deep representations $F=\left\{f_{i} \in \mathbb{R}^{d}\right\}_{i=1}^{N}$:
\begin{equation}
  S_{i j}^{F}= \, <f_{i}, f_{j}>
\end{equation}
where $<\cdot>$ represents dot product. 
After that, we use the feature similarity matrix $S^F$ and the spatial consistency matrix $\beta$ to calculate the similarity matrix $S$:
\begin{equation}
  S=S^{F} \otimes \beta
\end{equation}
where $\otimes$ represents element-wise product, and the spatial consistency matrix $\beta$ has been calculated before. 

Second, we set a threshold $\tau_{S}$ and use it to binarize the similarity matrix $S$ to a binary similarity matrix $\hat{S}$. 
When the $i$-th correspondence and the $j$-th correspondence satisfy sufficient spatial consistency and similarity in the feature space at the same time, $\hat{S}_{i j}=1$, otherwise $\hat{S}_{i j}=0$. 
Now, we can treat the input correspondences as the nodes of a graph, and $\hat{S}$ as the graph's adjacency matrix. 
The inlier sets can be regarded as subgraphs with a certain scale while outliers appear as isolated points or small-scale subgraphs. 

Finally, we sum the rows of the binary similarity matrix $\hat{S}$, and select the correspondences with row-sum values larger than a threshold $\tau_N$. 
As illustrated in Figure \ref{fig2}(a), our pruning method can effectively remove most or even all outliers. 
Our experiments also demonstrate that the pruning step is crucial for our method. 
We also provide details about how to select the thresholds $\tau_S$ and $\tau_N$ and ablation experiments in Supplementary Material. 

\subsection{Clustering and Transformations Estimation}\label{sec4.3}
After pruning, we obtain a clean set of correspondences, which is almost free of outliers. 
The next step is to divide these correspondences into multiple subsets belonging to different instances and estimate the final rigid transformations for all the instances. 
The correspondence division can be considered as a clustering problem, and the number of instances should be equal to the number of clusters. 
Here, we use a spectral clustering algorithm \cite{ref44,ref45}, which can determine the number of clusters automatically. 
The algorithm consists of the following steps:

Step 1: Recompute the binary similarity matrix $\hat{S}_n$ using the feature similarity and spatial consistency of the remaining correspondences, where $n$ represents the number of correspondences after pruning;

Step 2: Calculate the normalized Laplacian matrix $\hat{L}_n$ of the matrix $\hat{S}_n$, and calculate the eigenvalues $\lambda_{1}\left(\hat{L}_{n}\right) \leq \lambda_{2}\left(\hat{L}_{n}\right) \leq \ldots \leq \lambda_{n}\left(\hat{L}_{n}\right)$ of the normalized Laplacian matrix;

Step 3: Determine the number of clusters $M$ by the following formula: 
\begin{equation}
  \label{eq5}
  M=\underset{k}{\arg \max }\left\{\lambda_{k+1}\left(\hat{L}_{n}\right)-\lambda_{k}\left(\hat{L}_{n}\right)\right\}
\end{equation}

Step 4: Apply spectral clustering \cite{ref44} with $M$ clusters.

Since the deep representations have a good distribution in the feature space, the binary similarity matrix $\hat{S}_n$ conforms to the ideal matrix defined in \cite{ref45}. 
Therefore, the number of clusters determined by equation \ref{eq5} is reliable, which is proved in \cite{ref45}. 

After the above steps, the remaining correspondences are assigned to different clusters as shown in Figure \ref{fig2}(a), and then we can use a solver such as RANSAC \cite{ref9} to estimate the rigid transformation of each instance. 
Because our pruning and clustering strategy are very efficient, the inlier rate of each instance is high, and only a few dozen RANSAC iterations are enough to achieve outstanding performance. 

\section{Experiment}
We conduct experiments on both synthetic and real datasets and compare our PointCLM to state-of-the-art methods \cite{ref4,ref14,ref42,ref43}. 
The following sections are organized as follows. 
First, we illustrate our experimental settings including our implementation, datasets, competitors and evaluation metrics in section \ref{sec5.1}. 
Next, we conduct experiments on the synthetic and the real datasets in section \ref{sec5.2} and \ref{sec5.3}, respectively. 
We further conduct comprehensive ablation studies in section \ref{sec5.4} to illustrate the efficiency of our PointCLM and the importance of each component. 
More details are provided in the supplementary material, including the construction of the datasets, more qualitative evaluation and some experiments for hyperparameter choices. 

\subsection{Experimental Settings}\label{sec5.1}
\textbf{Implementation:} We implement our network in Pytorch \cite{ref47} and implement spectral clustering with sklearn \cite{ref48}. 
We randomly downsample the input correspondences to 1000. 
The distance parameter $\sigma_d$ is set to 0.05 for the synthetic dataset and 0.1 for the real dataset. 
We set the dimension $d$ of the deep representation to 128. 
Threshold $\tau_S$ is set to 0.85. 
Threshold $\tau_N$ is set to 10 for the synthetic dataset and 20 for the real dataset. 
The margins $m_p$ and $m_n$ are set to 0.1 and 1.4, respectively. 
Our batchsize is set to 16. 
We optimize the network using the ADAM optimizer with an initial learning rate of 0.01 and train the network for 15K iterations. 
All the experiments are conducted on a single RTX1080Ti graphic card with Intel Core i7-7800X CPU. 

\noindent
\textbf{Datasets:} We conduct experiments on both synthetic and real datasets. 
Our synthetic dataset is constructed from ModelNet40 \cite{ref49}, which consists of 12311 meshed CAD models from 40 categories. 
To construct our synthetic dataset, for each model, we uniformly downsample 1024 points from it to form the source point cloud, and then rotate and translate it 5-10 times repeatedly to generate multiple instances. 
The instances are mixed with noise points to form the target point cloud as shown in Figure \ref{fig3}. 
The rotation along each axis is uniformly sampled in [0,180$^{\circ}$] and the translation is in [0,5]. 
Our input correspondences for synthetic dataset are randomly generated by mixing the ground truth correspondences with outliers. 
We control the inlier ratio per instance at approximate 2\%. 
We generate 12311 such synthetic source-target point cloud pairs using 12311 models. 
We use 9843 pairs for training and 2468 pairs for testing. 
We randomly set aside 10\% pairs in the training set for validation.  

Our real dataset is Scan2CAD \cite{ref50}, which is constructed using ShapeNet \cite{ref51} and ScanNet \cite{ref52}. 
This dataset uses the CAD models in ShapeNet to replace the point clouds in the real scanned scene and provides accurate annotations, including models’ categories, rotations and translations, etc. 
The dataset provides 1506 annotated scenes, and each scene contains at least one class of instances. 
Therefore, we make full use of these annotations and split the scenes containing multiple kinds of instances into multiple source-target point cloud pairs for multi-instance registration. 
In this way, we get 2184 pairs of point clouds, most of which contain 2-5 instances of the same class in the target point clouds. 
We divide the samples into training set, validation set and test set according to the ratio of 7:1:2. We use fine-tuned FCGF \cite{ref23} to produce local features and generate putative correspondences by feature matching. 
More details about our dataset construction are provided in the supplementary material. 

\noindent
\textbf{Competitors:} We compare our PointCLM with four state-of-the-art methods, including T-linkage \cite{ref14}, RansaCov \cite{ref43}, CONSAC \cite{ref42} and ECC \cite{ref4}. 
T-linkage is a typical  algorithm based on preference analysis, which samples a series of hypotheses and clusters the inputs based on the residuals of the hypotheses. 
RansaCov regards  multi-model fitting as a maximum coverage problem, which can be approximately solved in greedy strategy or using relaxed linear programming. 
Here we use the former strategy due to its effectiveness. 
CONSAC uses a deep network to guide the sampling process, and we train this network using the same training set as our network. 
ECC clusters the input correspondences based on spatial consistency and doesn’t need hypothesis sampling as above methods. 
For a fair comparison, we not only use the same input, but also fine-tune the above methods both on GPU and CPU, and choose the ones with the best performance for comparison. 

\noindent
\textbf{Evaluation metrics:} We first define rotation error $RE$ and translation error $TE$ as follow:
\begin{equation}
  R E(R)=\arccos \left(\frac{\operatorname{Tr}\left(R^{\mathrm{T}} R^{*}\right)-1}{2}\right), T E(t)=\left\|t-t^{*}\right\|_{2}
\end{equation}
where $R^{*}$ and $t^{*}$ are the ground truth rotation and translation. 
We consider the instances with both rotation error and translation error below the thresholds to be successfully registered and our thresholds for $RE$ and $TE$ are 15$^{\circ}$ and 0.1, respectively. 
Since both our method and the above comparing methods predict multiple rigid transformations, we use mean recall (MR), mean precision (MP) and mean F1 Score (MF) as evaluation metric. 
For a pair of source point cloud and target point cloud, we define instance recall as $\frac{n_{ {success }}^{ {pred }}}{M^{ {gt }}}$ and instance precision as $\frac{n_{ {success }}^{ {pred }}}{M^{ {pred }}}$, where $n_{success}^{pred}$ denotes the number of successfully registered instances in prediction, $M^{gt}$ denotes the ground truth number of instances, and $M^{pred}$ denotes the number of predicted transformations. 
The instance F1 Score is the harmonic mean of the instance precision and instance recall. 
We calculate the instance precision, instance recall, and instance F1 Score of each sample in test set and average them to obtain our final evaluation metrics, the mean recall (MR), the mean precision (MP), and the mean F1 Score (MF). 

\subsection{Experiment on Synthetic Dataset}\label{sec5.2}
We first compare our method with other competitors on the synthetic dataset and the results are show in Table \ref{table1}. 
The sampling-based methods such as T-Linkage, RansaCov, and CONSAC do not achieve good performance due to extremely low inlier ratio. 
Benefiting from spatial consistency, ECC performs effectively. 
Nevertheless, our PointCLM surpasses the second best method ECC by a large margin in all evaluation metrics.

\setlength{\tabcolsep}{8pt}
\begin{table}[ht]
  \centering
  \caption{Registration results on synthetic dataset.}
  \begin{tabular}{ccccc}
    \hline\noalign{\smallskip}
  & MR(\%)         & MP(\%)         & MF(\%)         & Time(s)       \\
  \noalign{\smallskip}
  \hline
  \noalign{\smallskip}
  T-linkage \cite{ref14} & 0.61           & 1.48           & 0.87           & 3.89          \\
  RansaCov \cite{ref43}  & 0.73           & 5.33           & 1.29           & 0.14          \\
  CONSAC \cite{ref42}    & 1.00           & 7.45           & 1.77           & 0.61          \\
  ECC \cite{ref4}        & 82.90          & 92.92          & 87.63          & 3.56          \\
  PointCLM              & \textbf{92.60} & \textbf{99.69} & \textbf{96.01} & \textbf{0.06} \\
  \hline
  \end{tabular}
  \label{table1}
\end{table}

\begin{figure}
  \centering
  \includegraphics[width=1.0\textwidth]{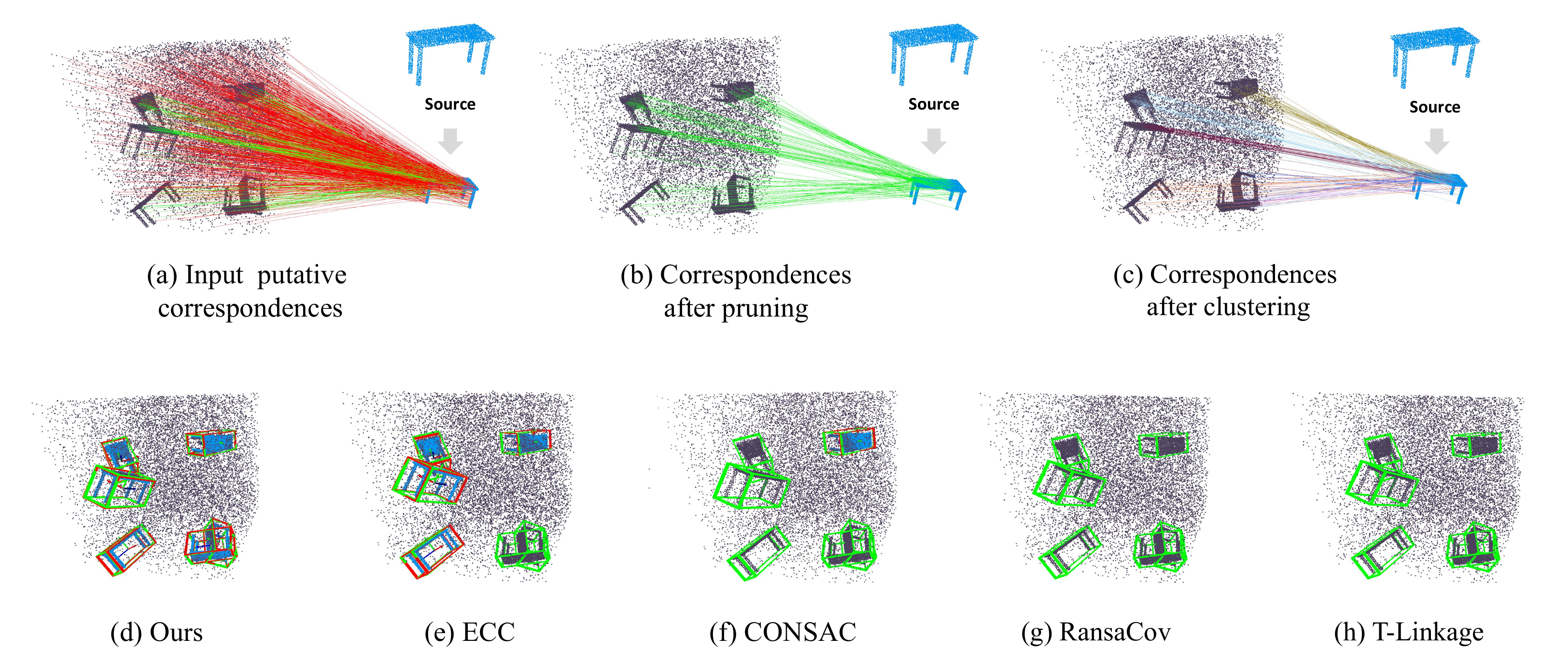}
  \caption{
    Results on synthetic dataset. 
    In (a) and (b), the green lines and red lines represent inlier correspondences and outlier correspondences, respectively. 
    In (c), the correspondences in each clusters after clustering are visualized in different colors. 
    In (d-f), the green bounding boxes represent the ground truth poses of instances in the target point cloud and the red bounding boxes represent predictions. 
    The transformed point clouds in the target point cloud are visualized in blue.
  }
  \label{fig3}
\end{figure}

We provide a set of visualizations to qualitatively evaluate our PointCLM and compare it with other competitors in Figure \ref{fig3}. 
The first row of Figure \ref{fig3} shows the input correspondences, and our pruning and clustering results. 
Figure \ref{fig3}(b) shows that our PointCLM surprisingly removes all outliers, and the remaining correspondences are well clustered as shown in Fig. \ref{fig3} (c). 
The second row of Figure \ref{fig3} shows the registration results of our proposed method and the competitors. 
It can be seen that both T-Linkage and RansaCov fail to register any instances. 
For the six instances in the target point cloud, CONSAC only registers one instance successfully. 
It is worth noting that although ECC register four instances successfully, but it fails to register the two tables in the lower right corner. 
This is because the two tables are mixed together and spatial consistency is not enough to distinguish them. 
However, our method successfully registers all instances. 

\setlength{\tabcolsep}{1.4pt}

\setlength{\tabcolsep}{8pt}
\begin{table}
  \centering
  \begin{center}
  \end{center}
  \caption{Registration results on real dataset.}
  \begin{tabular}{ccccc}
  \hline\noalign{\smallskip}
  & MR(\%)         & MP(\%)         & MF(\%)         & Time(s)       \\
  \noalign{\smallskip}
  \hline
  \noalign{\smallskip}
  T-linkage \cite{ref14} & 34.99          & 46.86          & 40.07          & 6.64          \\
  RansaCov \cite{ref43}  & 60.50          & 33.28          & 42.94          & \textbf{0.07} \\
  CONSAC \cite{ref42}    & 55.48          & 53.34          & 54.39          & 0.39          \\
  ECC \cite{ref4}       & 64.66          & 69.73          & 67.10          & 1.84          \\
  PointCLM              & \textbf{78.10} & \textbf{70.64} & \textbf{74.18} & 0.10        \\ 
  \hline
  \end{tabular}
  \label{table2}
\end{table}
\setlength{\tabcolsep}{1.4pt}

\subsection{Experiment on Real Dataset}\label{sec5.3}
We then compare our PointCLM with other competitors on Scan2CAD \cite{ref50}. 
As shown in Table \ref{table2}, our PointCLM outperforms all the competitors on all three evaluation metrics, MR, MP, and MF and it is also competitive in speed. 
The performance of ECC and PointCLM is lower than that on the synthetic dataset while the performance of the other methods is higher than that on the synthetic dataset. 
This is due to the change of the distribution of the instances' inlier ratio. 

\begin{figure}
  \centering
  \includegraphics[width=1.0\textwidth]{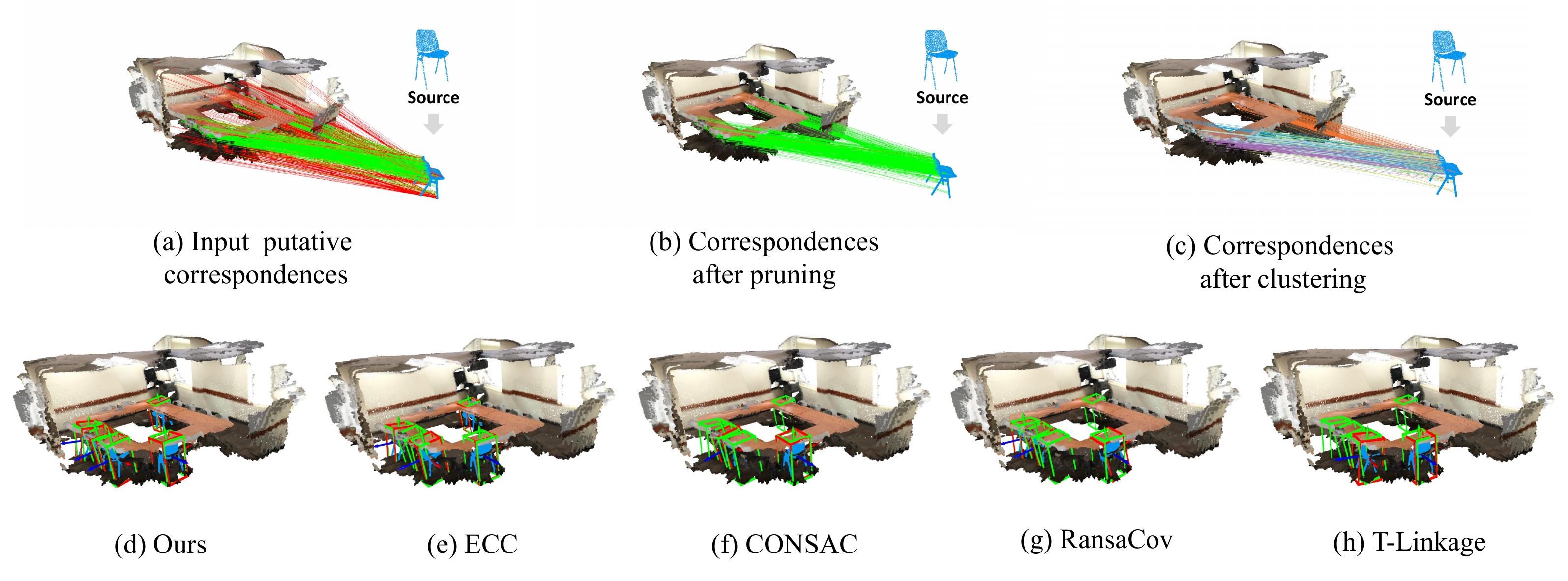}
  \caption{
    Results on real dataset. 
    In (a) and (b), the green lines and red lines represent inlier correspondences and outlier correspondences, respectively. 
    In (c), the correspondences in each clusters after clustering are visualized in different colors. 
    In (d-f), the green bounding boxes represent the ground truth poses of instances in the target point cloud and the red bounding boxes represent predictions. 
    The transformed point clouds in the target point cloud are visualized in blue.
  }
  \label{fig4}
\end{figure}

We also provide a set of visualizations to qualitatively evaluate our PointCLM and compare it with the other competitors. 
The first row of Figure \ref{fig4} shows the input correspondences, and our pruning and clustering results. 
Figure \ref{fig4}(b) shows that our method removes almost all outliers, ensuring that the following clustering performs efficiently as shown in Figure \ref{fig4}(c). 
For the five instances contained in the target point cloud, T-Linkage and CONSAC successfully register two instances, but one prediction of T-Linkage has large errors. 
RansaCov successfully registers three instances. 
Since three chairs in the target point cloud are close to each other, ECC does not successfully register all these instances. 
Our method successfully accomplishes the registration of all instances.

\subsection{Ablation Studies}\label{sec5.4}
In this section we provide comprehensive ablation studies to illustrate the effectiveness of each component. 
All our ablation studies are performed on Scan2CAD.

\noindent
\textbf{Ablation on deep representation:} To study the effectiveness of our adopted deep representations, we compare the performance of our framework with and without deep representation. 
The version without deep representation deletes Feature Extractor and set $S=\beta$, which relies only on spatial consistency for pruning and clustering. 
The comparison results are shown in Table \ref{table3}. 
It can be seen that both the accuracy and the speed metrics are improved by using the deep representations. 
Without deep representations, less correspondences are pruned and more correspondences need to be clustered, which increase the runtime of our method. 
Although the framework without deep representations has lower performance in MR, MP and MF, it is still a competitive baseline, which is suitable for the case without training data. 
This indicates the effectiveness of the pruning and clustering strategies proposed in this paper.

\begin{figure}
  \centering
  \includegraphics[width=0.8\textwidth]{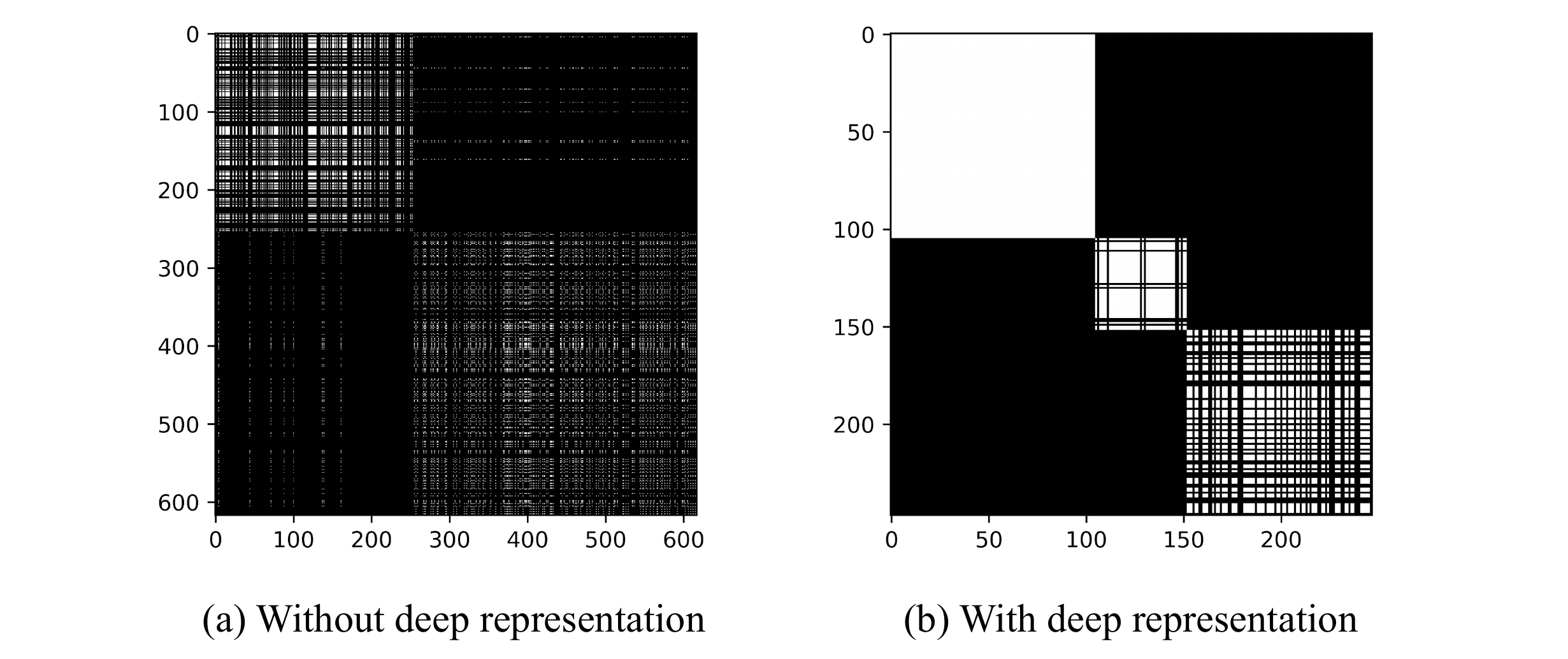}
  \caption{
    Visualization of clustering results without and with deep representation. 
    }
  \label{fig5}
\end{figure}

Additionally, we visualize the clustering results with and without deep representation in Figure \ref{fig5}. 
We select an example, whose target point cloud contains three instances. 
We first use a 3-dimensional one-hot vector to represent which instance a correspondence belongs to. 
Then we use these vectors to calculate similarity matrices and permute these matrices with the results of the clustering. 
It can be seen that the similarity matrix permuted by the framework with deep representation is much smaller than the one permuted by the framework without deep representation because more outliers are removed during pruning in the former case. 
More importantly, the matrix in Figure \ref{fig5}(b) shows three clear clusters, which correspond to the three instances. 
On the contrary, the matrix in Figure \ref{fig5}(a) shows two blocks, where the lower right block actually corresponds to two instances, which cannot be distinguished successfully without using the proposed deep representation. 

\setlength{\tabcolsep}{2pt}
\begin{table}
  \caption{
    Ablation study results on deep representation and pruning.
    }
  \centering
  \begin{tabular}{cccccc}
    \hline\noalign{\smallskip}
    Deep representatoin & Pruning & MR(\%)         & MP(\%)         & MF(\%)         & Time(s)       \\
  \noalign{\smallskip}
  \hline
  \noalign{\smallskip}
  & $\checkmark$       & 76.61          & 65.05          & 70.36          & 0.17          \\
  $\checkmark$                   &         & 62.23          & 32.77          & 42.93          & 1.24          \\
  $\checkmark$                & $\checkmark$     & \textbf{78.10} & \textbf{70.64} & \textbf{74.18} & \textbf{0.10}\\
  \hline
  \end{tabular}
  \label{table3}
\end{table}
\setlength{\tabcolsep}{1.4pt}

\noindent
\textbf{Ablation on pruning:} To quantitatively study the effectiveness of our pruning strategy, we simply remove the pruning strategy in our method, and then compare the performance before and after removal. 
Table. \ref{table3} shows that the performance of our method drops sharply if the pruning step is removed, which is due to the fact that the noisy binary similarity matrix does not conform to the ideal matrix defined in \cite{ref45} and spectral clustering cannot group the correspondences correctly. 
This result reveals that pruning is crucial for subsequent clustering. 

\noindent
\textbf{Ablation on RANSAC iterations:} In addition, we also test the performance of our model with different numbers of RANSAC iterations and the results are shown in Table \ref{table4}. 
It can be seen that fairly good results can be achieved with only five iterations, and the performances with 50 and 500 iterations are very close. 
These results indicate that the pruning and clustering steps greatly improve the inlier ratio of each instance, so reliable results can be estimated using only a small number of iterations.

\setlength{\tabcolsep}{12pt}
\begin{table}
  \caption{
    Influence of the number of RANSAC iterations.
  }
  \centering
  \begin{tabular}{cccc}
    \hline\noalign{\smallskip}
    RANSAC iterations & MR(\%)         & MP(\%)         & MF(\%)          \\
  \noalign{\smallskip}
  \hline
  \noalign{\smallskip}
  5   & 77.39          & 67.76          & 72.26              \\
  50  & 78.10          & \textbf{70.64} & 74.18                \\
  500 & \textbf{79.33} & \textbf{70.64} & \textbf{74.73}  \\  
  \hline
  \end{tabular}
  \label{table4}
\end{table}
\setlength{\tabcolsep}{5pt}

\noindent
\textbf{Ablation on hardest negative pairs:} 
Exhausting the negative pairs to find the hardest ones for each minibatch incurs larger computational overhead, but our ablation shows it worthwhile. 
We compare it with the same model trained using randomly selected negative pairs. 
The results of using the hardest negative pairs as in the proposed method and using randomly selected negative pairs are shown in Table \ref{table5}. 
All accuracy metrics are improved by using hardest negative pairs in contrastive learning. 
By comparing Table \ref{table5} and Table \ref{table3}, we can find that using random negative pairs in contrastive representative learning achieves better results than not using the deep representation, but the margin is small. 
In addition, we collect the average cosine similarity within positive pairs and top-K hardest negative pairs in the feature space in Table \ref{table5}. 
The result shows the representation trained using hardest negative pairs are better separated in the feature space, which is more discriminative.

\setlength{\tabcolsep}{2pt}
\begin{table}
  \caption{
    Ablation experiment results on training with hardest negative pairs. 
  }
  \centering
  \begin{tabular}{ccccccc}
    \hline\noalign{\smallskip}
    & MR(\%) & MP(\%) & MF(\%) & Positive(\%) & Top-1(\%) & Top-10(\%) \\
    \noalign{\smallskip}
    \hline
    \noalign{\smallskip}
    Random  & 76.85  & 67.50  & 71.87  & 95.43 & 96.91 & 91.37  \\
    Hardest & \textbf{78.10} & \textbf{70.64} & \textbf{74.18} & 83.96 & 61.32 & 49.65 \\  
    \hline
  \end{tabular}
  \label{table5}
\end{table}
\setlength{\tabcolsep}{1.4pt}


\section{Conclusion}
In this paper, we propose a novel framework to address the multi-instance point cloud registration problem. 
We use contrastive learning to learn well-distributed deep representations for input correspondences, based on which we develop a pruning and a clustering strategy to remove outlier correspondences efficiently and assign the remaining inlier correspondences to correct instances. 
Then, the transformation from the source point cloud to each instance can be easily estimated. 
Extensive experiments on both synthetic and real datasets demonstrate the effectiveness of our framework and its superiority over existing solutions. 
We think the proposed representation learning and the outlier pruning strategy has the potential to be used in pairwise point cloud registration. 


%
%
\bibliographystyle{splncs04}
\bibliography{egbib}

\clearpage
\appendix
\section{Dataset Construction}
Section \ref{sec5.1} has explained how to generate the source point cloud and the target point cloud on Scan2CAD \cite{ref50}. 
Here, we mainly focus on how to generate the putative correspondences by  feature matching. 
We used the FCGF \cite{ref23} as feature extractor, whose parameter setting is shown in Table \ref{table6}. 
The FCGF network is pretrained on 3DMatch dataset and then fine-tuned with parameter setting in Table \ref{table7}. 
The fine-tuned FCGF extracts L2-normalized local feature $F_{X}^{ {local }}=\left\{f_{x_{i}}^{ {local }} \in R^{32}\right\}_{i=1}^{|X|}$ for the source point cloud $X$ and $F_{Y}^{{local }}=\left\{f_{y_{i}}^{{local }} \in R^{32}\right\}_{i=1}^{|Y|}$ for the target point cloud $Y$, where $|X|$ and $|Y|$ denote the number of points in the source point cloud and the target point cloud, respectively. 
Given each point $y_j \in Y$ in the target point cloud, we find the point $x_i \in X$ in the source point cloud satisfying $i=\underset{i}{\arg \max }<f_{y_{j}}^{{local }}, f_{x_{i}}^{{local }}>$ to build correspondence $(x_i,y_j)$, where $<f_{y_{j}}^{{local }}, f_{x_{i}}^{{local }}>$ is the cosine similarity between two point features. 
In this way, we obtain ${|Y|}$ correspndences and we define the cosine similarity $<f_{x_{i}}^{{local }}, f_{y_{j}}^{{local }}>$ as the saliency score of correspondence $(x_i,y_j)$. 
We select $K$ correspondences with largest saliency scores and then randomly downsample these $K$ correspondences to $N$ correspondences as the input putative correspondences. 
Here, we set $K$ as 10000 to make input correspondences cover as many instances as possible. 
$N$ is set to 1000, which has been stated  in section \ref{sec5.1}.
\vspace{-1em}

\setlength{\tabcolsep}{12pt}
\begin{table}
  \caption{
    Parameter setting and pretraining of the FCGF network.  
  }
  \centering
  \begin{tabular}{c|c}
    \hline
    Model & RESUNETBN2C   \\
    Downsampling voxel size & 2.5cm (0.025) \\
  Feature dimension       & 32            \\
  Pretrained dataset      & 3DMatch       \\
  Normalized feature      & True         \\
    \hline
  \end{tabular}
  \label{table6}
\end{table}
\setlength{\tabcolsep}{1.4pt}
\vspace{-2em}

\setlength{\tabcolsep}{12pt}
\begin{table}
  \caption{
    Parameter setting in fine-tuning the FCGF network.
  }
  \centering
  \begin{tabular}{c|c}
    \hline
    Batch size    & 4    \\
  Learning rate & $10^{-3}$ \\
  Epoch         & 20   \\
  Optimizer     & SGD         \\
    \hline
  \end{tabular}
  \label{table7}
\end{table}
\setlength{\tabcolsep}{1.4pt}
\vspace{-0.5em}

\begin{figure}
  \centering
  \includegraphics[width=0.8\textwidth]{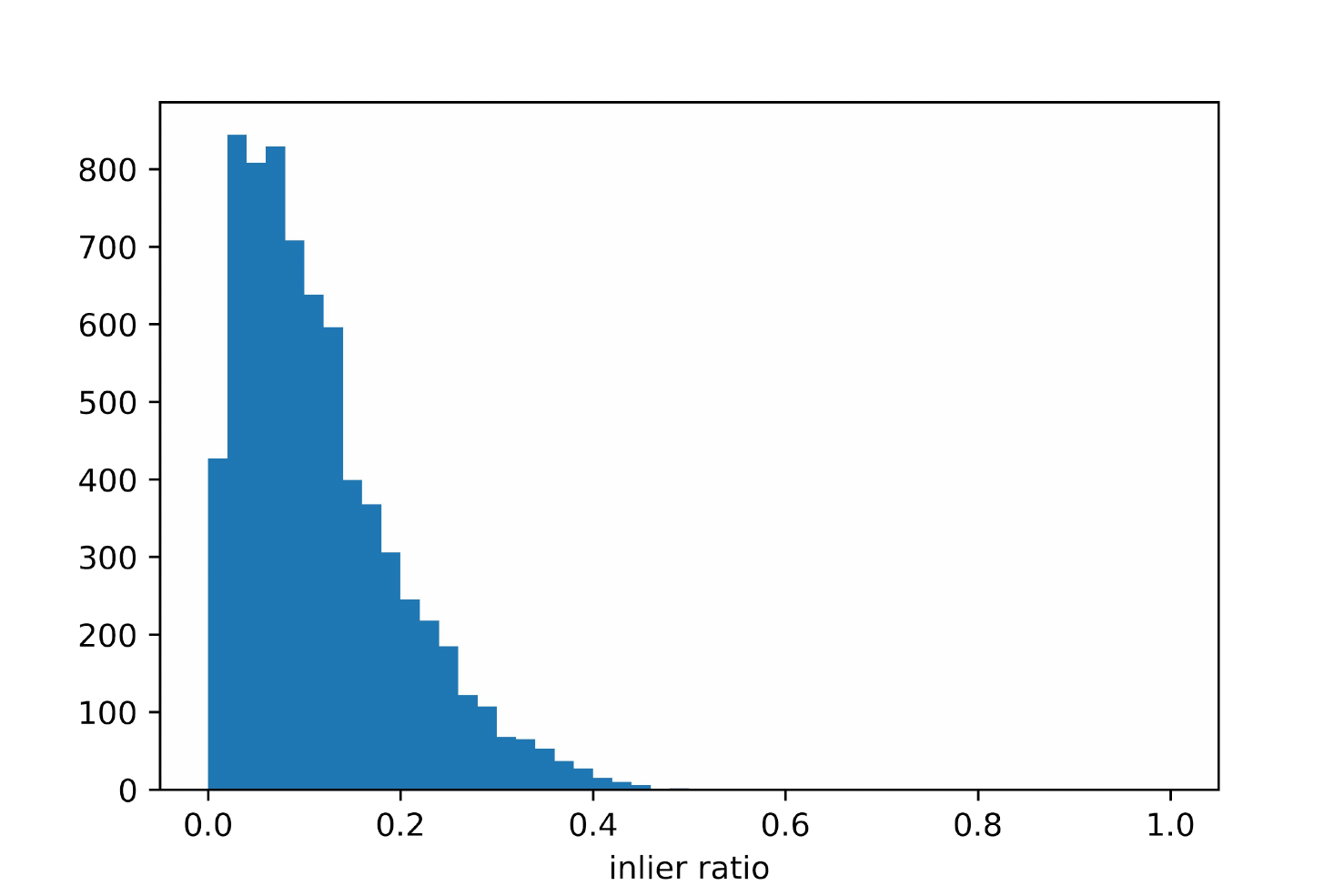}
  \caption{
  The histogram of per-instance inlier ratio on Scan2CAD. 
  }
  \label{fig6}
\end{figure}

The per-instance inlier ratio of the input putative correspondences for the Scan2CAD dataset is show in Figure \ref{fig6}. 
We can see that most instances have an inlier rate of less than 10\%, and many are less than 2\%.


\section{Hyperparameters Choice}
The threshold $\tau_S$ and $\tau_N$ are key hyperparameters of our proposed pruning strategy, which are used to binarize the compatibility between correspondences and identity the inlier sets. 
We first evaluate the performance of our framework with different thresholds $\tau_S$ and the results are show in Table \ref{table8}. 
Our framework is robust to the choices of threshold $\tau_S$, and the performance of our framework is significantly superior to existing methods with all different values of it. 
This is because our correspondences are well separable in the feature space. 
We calculate the average cosine similarity between correspondences in the feature space and the results are shown in Table \ref{table9}. 
In Table \ref{table9}, Positive denotes the average cosine similarity between anchor correspondences and their positive samples in the feature space. 
Top-K denotes the average cosine similarity between anchor correspondences and their top-K hardest samples in the feature space. 
There exists a large margin between Positive and Top-K, which indicates the correspondences are well separated in the feature space. 
The choice of threshold $\tau_S$ is a tradeoff, which slightly affected the performance of our framework. 
If we choose a smaller $\tau_S$, fewer correspondences will be pruned, which may improve the recall but decrease the precision.

\setlength{\tabcolsep}{12pt}
\begin{table}
  \caption{
    Performance of our PointCLM when varying the threshold $\tau_S$. 
  }
  \centering
  \begin{tabular}{cccc}
    \hline \noalign{\smallskip}
      & MR(\%) & MP(\%) & MF(\%) \\
    \noalign{\smallskip}
    \hline
    \noalign{\smallskip}
    0.70 & 78.54 & 68.29 & 73.06 \\
  0.75 & \textbf{79.35} & 68.81  & 73.71  \\
  0.80 & 78.81 & 68.64 & 73.37 \\
  0.85 & 78.10  & \textbf{70.64} & \textbf{74.18} \\
  0.90 & 74.06 & \textbf{70.64} & 72.31  \\
   \hline
  \end{tabular}
  \label{table8}
\end{table}
\setlength{\tabcolsep}{1.4pt}

\setlength{\tabcolsep}{12pt}
\begin{table}
  \caption{
    Average cosine similarity within positive pairs and top-K hardest negative pairs in the feature space.
  }
  \centering
  \begin{tabular}{ccccc}
    \hline \noalign{\smallskip}
    Positive & Top-1 (\%) & Top-5 (\%) & Top-10 (\%) & Top-15 (\%) \\
    \noalign{\smallskip}
    \hline
    \noalign{\smallskip}
    83.96    & 61.32      & 52.90      & 49.65       & 47.59    \\
   \hline
  \end{tabular}
  \label{table9}
\end{table}
\setlength{\tabcolsep}{1.4pt}

Then we evaluate the performance of our framework with different $\tau_N$ and the results are shown in Table \ref{table10}. 
Again, our framework outperforms all existing methods with all different choices of $\tau_N$, though it has some influence on each metric. 
The choice of the threshold $\tau_N$ is also a trade-off. 
If we choose a smaller $\tau_N$, some outlier clusters may be considered as instances and some instances with small inlier ratios may be registered successfully, which results in a lower precision and a higher recall. 
In practice, fine-tuning of the parameters can be performed on a validation set.

\setlength{\tabcolsep}{12pt}
\begin{table}
  \caption{
    Performance of our PointCLM when varying the threshold $\tau_N$. 
  }
  \centering
  \begin{tabular}{cccc}
    \hline \noalign{\smallskip}
    & MR(\%)         & MP(\%)         & MF(\%)         \\
    \noalign{\smallskip}
    \hline
    \noalign{\smallskip}
    10 & \textbf{80.60} & 61.44          & 69.73          \\
  15 & 78.74          & 66.82          & 72.29          \\
  20 & 78.10          & 70.64          & \textbf{74.18} \\
  30 & 76.56          & \textbf{71.28} & 73.83    \\
   \hline
  \end{tabular}
  \label{table10}
\end{table}
\setlength{\tabcolsep}{1.4pt}



\end{document}